\documentclass[journal]{IEEEtran}

\ifCLASSINFOpdf
\else
   \usepackage[dvips]{graphicx}
\fi
\usepackage{url}

\usepackage{graphicx}
\usepackage{booktabs}
\usepackage{amsmath}
\usepackage{xcolor}
\usepackage{makecell}
\usepackage{multirow}
\usepackage{pifont}

\begin{document}

\title{Progressive Motion Context Refine Network for Efficient Video Frame Interpolation}

\author{Lingtong Kong, \, Jinfeng Liu, \, Jie Yang
\thanks{The authors are with the Institute of Image Processing and Pattern Recognition, Shanghai Jiao Tong University, Shanghai 200240, China (e-mail: ltkong@sjtu.edu.cn, ljf19991226@sjtu.edu.cn, jieyang@sjtu.edu.cn).}}

\maketitle

\begin{abstract}
Recently, flow-based frame interpolation methods have achieved great success by first modeling optical flow between target and input frames, and then building synthesis network for target frame generation. However, above cascaded architecture can lead to large model size and inference delay, hindering them from mobile and real-time applications. To solve this problem, we propose a novel Progressive Motion Context Refine Network (PMCRNet) to predict motion fields and image context jointly for higher efficiency. Different from others that directly synthesize target frame from deep feature, we explore to simplify frame interpolation task by borrowing existing texture from adjacent input frames, which means that decoder in each pyramid level of our PMCRNet only needs to update easier intermediate optical flow, occlusion merge mask and image residual. Moreover, we introduce a new annealed multi-scale reconstruction loss to better guide the learning process of this efficient PMCRNet. Experiments on multiple benchmarks show that proposed approaches not only achieve favorable quantitative and qualitative results but also reduces current model size and running time significantly.
\end{abstract}


\IEEEpeerreviewmaketitle

\section{Introduction}
\IEEEPARstart{V}{ideo} frame interpolation (VFI) is an important low-level computer vision task, and has gathered much attention from both industry and academia, due to its widespread applications to slow motion generation~\cite{8579036}, novel view synthesis~\cite{Zhou_2016} and video compression~\cite{Wu_2018_ECCV}. With the advent of convolutional neural networks (CNNs), frame interpolation accuracy has been greatly improved by deep-based methods~\cite{8099727,8578281,8954114,choi2020cain,Lee_2020_CVPR,Liu_2022_ICIP}. However, the execution efficiency for frame interpolation task has not been fully studied, which restricts lots of mobile and real-time applications.

Existing deep-based VFI algorithms can be roughly classified into flow-based~\cite{8237740,8579036,8578281,8954114,Niklaus_2020_CVPR,BMBC} and kernel-based~\cite{8099727,Cheng_2020,choi2020cain,Lee_2020_CVPR,Gui_2020_CVPR} methods, where the first one contains a separate optical flow estimation and modeling stage, while, the latter one unifies motion perception and image generation into a series of adaptive convolution operations. Different VFI paradigms have their own advantages and disadvantages. To be specific, kernel-based methods can generate complex texture thanks to the strong representation ability of adaptive convolution~\cite{8099727,8237299}, however, their prediction tend to be blurry when scene motion is large~\cite{choi2020cain,Lee_2020_CVPR}. On the other hand, flow-based methods can synthesize target frame with clear dynamic texture in diverse motion cases~\cite{8578281,Niklaus_2020_CVPR}, while, their cascaded encoder-decoder networks usually contain large model size and suffer from slow running speed~\cite{8954114,BMBC}. Concretely, the hybrid methods of DAIN~\cite{8954114} and MEMC-Net~\cite{8840983} contains more than 70 M and 24 M parameters, respectively. Moreover, some cascaded flow-based VFI architectures, such as SoftSplat~\cite{Niklaus_2020_CVPR} and BMBC~\cite{BMBC}, run less than 10 fps on a common high-end GPU. Both of these defects have hindered them from mobile and real-time applications.

Recently, there are several methods that explore to improve efficiency for frame interpolation. For example, RIFE~\cite{huang2021rife} proposes an IFNet to directly estimate intermediate optical flow in real-time, and then builds a simple synthesis network for target frame generation. However, it separates flow estimation and context synthesis into different encoder-decoders, containing relatively large parameters. Then, CDFI~\cite{ding2021cdfi} presents a compression-driven network design for frame interpolation, where a channel pruning stage compresses the baseline AdaCoF~\cite{Lee_2020_CVPR} network, and then a pyramid synthesis compensation stage improves on the previous one. Although, it has reduced number of parameters obviously, their complex VFI architecture design has led to much more inference delay. Later on, CAIN-SD~\cite{Choi_2021_ICCV} puts up a dynamic architecture that first calculates the approximate motion magnitude of each image patch and then uses them to decide the depth of the model and the scale of the input. Compared with the baseline CAIN~\cite{choi2020cain}, it can reduce much computation on high-resolution videos. Whereas, this framework can not deal with large motion cases well and the interpolated frames tend to have sudden change at patch edges. Recently, FILM~\cite{reda2022film} builds a single unified network distinguished by a shared multi-scale feature extractor for both motion estimation and frame synthesis. However, they directly generate target frame from deep feature, which suffers from large computation cost and blur artifacts.

\begin{figure*}[t]
	\centering
	\includegraphics[width=\linewidth]{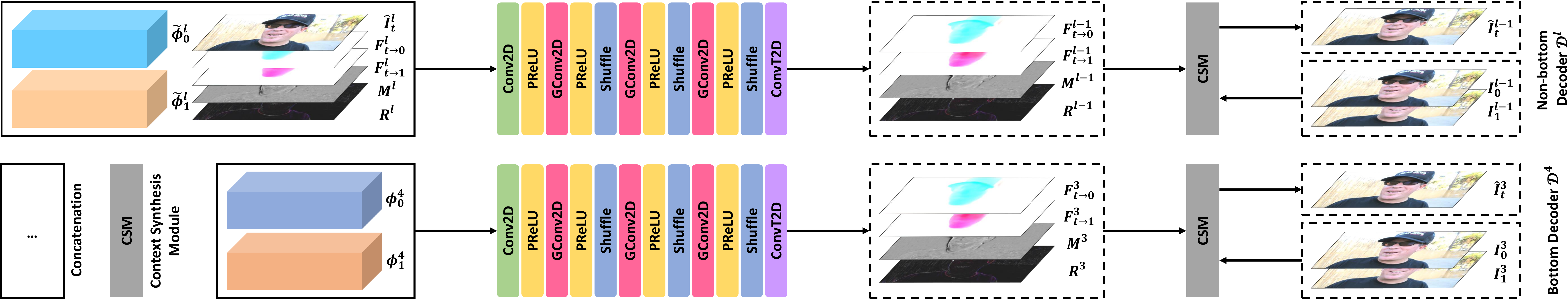}
	\caption{Structure details of multiple decoder sub-networks in proposed PMCRNet. `GConv2D', `Shuffle' and `ConvT2D' represent group convolution, channel shuffle operation~\cite{Zhang_2018_CVPR,Kong_2021_ICRA} and transposed convolution. `CSM' stands for proposed context synthesis module. $\mathcal{D}^{l}$ ($l \in \{1, 2, 3\}$) means non-bottom decoders.}
	\label{fig:1}
\end{figure*}

To solve above problems, in this letter, we propose a novel single encoder-decoder based progressive motion context refine network, termed PMCRNet, for efficient video frame interpolation. It first extracts pyramid features from two input frames, and then estimates intermediate optical flow and image context in a coarse-to-fine manner, which can deal with the large displacement challenge. Different from others that directly generate target frame from deep feature~\cite{Lee_2020_CVPR,Gui_2020_CVPR,Niklaus_2020_CVPR,BMBC,reda2022film}, we simplify the frame synthesis in each scale by borrowing existing texture from adjacent input image pyramids, which means that each decoder only needs to refine easier intermediate optical flow, occlusion merge mask and image residual. Corresponding to this progressive refinement architecture, we further introduce a new annealed multi-scale reconstruction loss to provide better learning guidance, where it does not need additional pre-trained optical flow network, and thus enjoys higher training and inference efficiency. Experiments on multiple VFI benchmarks verify the effectiveness of proposed contributions.

\section{Method}
In this section, we first introduce the network structure of proposed PMCRNet, and then present the loss function.

\subsection{Pyramid Encoder}
Given two input frames $I_0, I_1$, our goal is to estimate the intermediate frame $I_t$, where $t=0.5$. In order to deal with the large displacement challenge in VFI, we propose to predict intermediate frame in a progressive refinement manner. First, to acquire a hierarchical representation of input frames, a pyramid encoder network $\mathcal{E}$ is employed to extract multi-scale pyramid features. Specifically, encoder $\mathcal{E}$ consists four convolution blocks to extract four levels of pyramid features, where each block is made up of one stride 2 convolution followed by one stride 1 convolution. A PReLU activation~\cite{7410480} is next to each convolution layer. As the spatial size is gradually compressed, feature channels are correspondingly expanded to 48, 96, 144 and 192, yielding pyramid features $\phi_{0}^{l}, \phi_{1}^{l}$ in level $l$ ($l \in \{1, 2, 3, 4\}$) for $I_0$ and $I_1$, separately.

\begin{figure}[t]
	\centering
	\includegraphics[width=\linewidth]{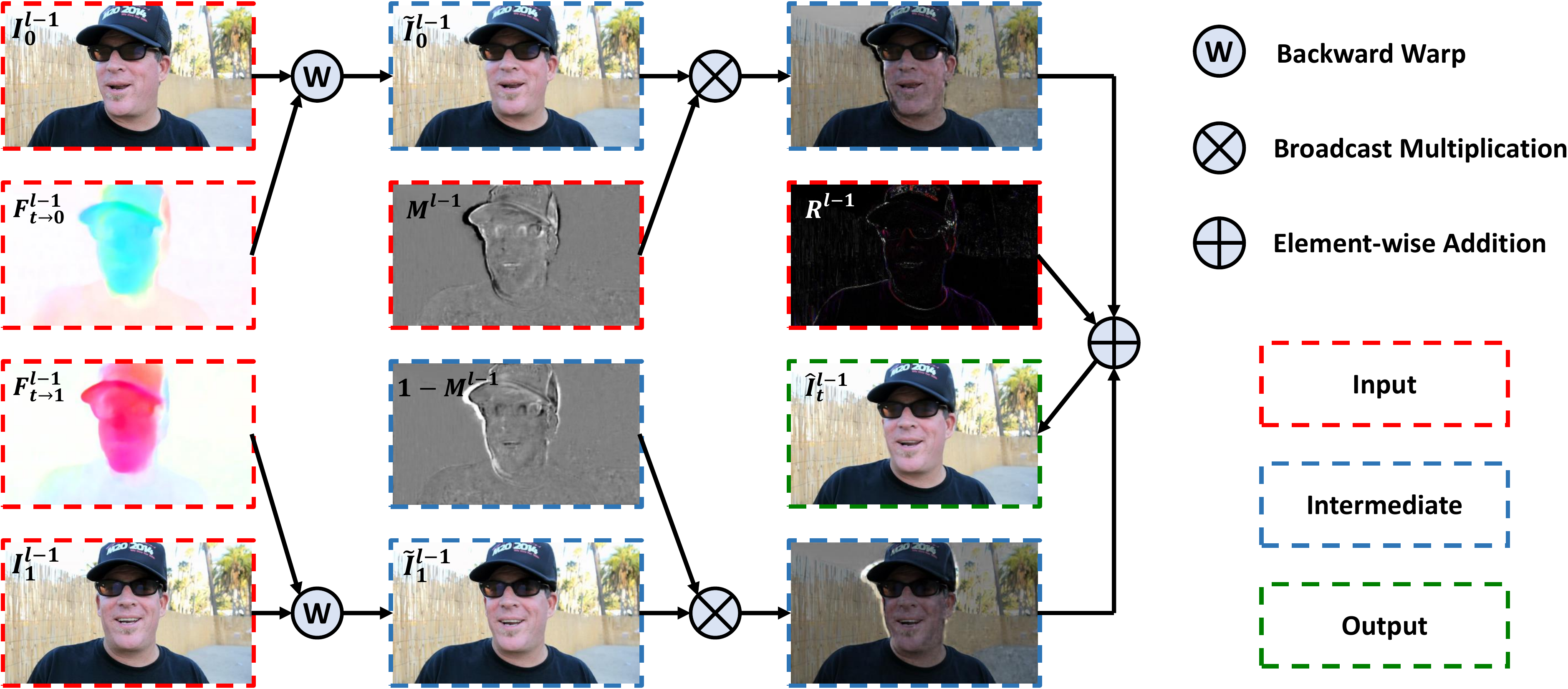}
	\caption{Structure details of context synthesis module (CSM) in decoder $\mathcal{D}^{l}$.}
	\label{fig:2}
\end{figure}

\subsection{Joint Refinement Decoder}
Given above hierarchical feature representation, different from other flow-based frame interpolation methods~\cite{8579036,8578281,8954114,Niklaus_2020_CVPR,BMBC,huang2021rife} that take cascaded deep architecture, proposed PMCRNet adopts coarse-to-fine decoders, \textit{i.e.}, $\mathcal{D}^{4}, \mathcal{D}^{3}, \mathcal{D}^{2}, \mathcal{D}^{1}$, to jointly refine intermediate optical flow $F_{t\rightarrow0}^{l}, F_{t\rightarrow1}^{l}$ and target frame context $\hat{I}_{t}^{l}$, \textit{i.e.}, predicted target frame in level $l$, for higher efficiency. Fig.~\ref{fig:1} depicts structure details of decoders in our PMCRNet, where the bottom decoder $\mathcal{D}^{4}$ is different from other non-bottom decoders only in the input part. Following the success of pyramid warping strategy in PWC-Net~\cite{8579029}, we use previously estimated intermediate flow to backward warp pyramid features to reduce motion displacement. On the other hand, warped pyramid features are aligned to target position which can also provide useful context information. In summary, input and output among two types of decoders can be written by
\begin{align}
	[F_{t\rightarrow0}^{3}, F_{t\rightarrow1}^{3}, M^{3}, R^{3}] & = \mathcal{D}^{4}([\phi_{0}^{4}, \phi_{1}^{4}]), \label{eq:1}\\
	[F_{t\rightarrow0}^{l-1}, F_{t\rightarrow1}^{l-1}, M^{l-1}, R^{l-1}] & = \mathcal{D}^{l}([\tilde{\phi}_{0}^{l}, \tilde{\phi}_{1}^{l}, F_{t\rightarrow0}^{l}, F_{t\rightarrow1}^{l}, \nonumber \\ M^{l}, R^{l}, \hat{I}_{t}^{l}]) \, , \, l \in \{1, 2, 3\}, \label{eq:2}
\end{align}
where $\tilde{\phi}_{0}^{l} = w(\phi_{0}^{l}, F_{t\rightarrow0}^{l})$, $\tilde{\phi}_{1}^{l} = w(\phi_{1}^{l}, F_{t\rightarrow1}^{l})$ and $w$ stands for backward warping. $M^{l-1}, R^{l-1}$ will be described later. To be specific, each $\mathcal{D}^{l}$ includes four convolution layers and one transposed convolution layer, with stride 1 and 1/2, respectively. Each convolution layer follows a PReLU activation, and the last three convolution layers are group convolution followed by channel shuffle operation for better efficiency, which has been verified by FastFlowNet~\cite{Kong_2021_ICRA} in optical flow.

\begin{table*}[t]
	\centering
	\renewcommand{\arraystretch}{0.9}
	\caption{Quantitative comparison with state-of-the-art frame interpolation methods that do \textbf{not} use additional flow knowledge on the Vimeo90K, UCF101 and Middlebury datasets. For each item, the best result is colored \textcolor{red}{\textbf{red}} and the second best is colored \textcolor{blue}{\textbf{blue}}.}
	\label{tab:1}
	{\footnotesize
		\setlength\tabcolsep{11.2pt}
		\begin{tabular}{lcccccccccc}
			\toprule
			\multirow{2}[2]{*}{Method} & \multirow{2}[2]{*}{\makecell{Training \\ Dataset}} & \multicolumn{2}{c}{Vimeo90K} & \multicolumn{2}{c}{UCF101} & \multicolumn{3}{c}{Middlebury} & \multirow{2}[2]{*}{\makecell{Time \\ (s)}} & \multirow{2}[2]{*}{\makecell{Params \\ (M)}} \\
			\cmidrule(lr){3-4} \cmidrule(lr){5-6} \cmidrule(lr){7-9}
			& & PSNR & SSIM & PSNR & SSIM & PSNR & SSIM & IE & & \\
			\midrule
			DVF~\cite{8237740} & UCF101 & 31.54 & 0.946 & 34.12 & 0.963 & - & - & 4.04 & 0.078 & \textcolor{blue}{\textbf{1.6}} \\
			SepConv~\cite{8237299} & proprietary & 33.79 & 0.970 & 34.78 & 0.967 & 35.73 & 0.959 & 2.27 & 0.091 & 21.6 \\
			SuperSlomo~\cite{8579036} & Adobe240 & 33.15 & 0.966 & 34.75 & \textcolor{blue}{\textbf{0.968}} & - & - & 2.28 & 0.059 & 19.8 \\
			CyclicGen~\cite{liu2019cyclicgen} & UCF101 & 32.10 & 0.949 & 35.11 & \textcolor{blue}{\textbf{0.968}} & - & - & 2.86 & - & 3.1 \\
			ToFlow~\cite{xue2019video} & Vimeo90K & 33.73 & 0.968 & 34.58 & 0.967 & 35.29 & 0.956 & 2.15 & 0.076 & \textcolor{red}{\textbf{1.4}} \\
			CAIN~\cite{choi2020cain} & Vimeo90K & 34.65 & 0.973 & 34.98 & \textcolor{red}{\textbf{0.969}} & 35.11 & 0.951 & 2.28 & 0.034 & 42.8 \\
			AdaCoF~\cite{Lee_2020_CVPR} & Vimeo90K & 34.27 & 0.971 & 34.91 & \textcolor{blue}{\textbf{0.968}} & 35.72 & \textcolor{blue}{\textbf{0.978}} & 2.31 & \textcolor{blue}{\textbf{0.031}} & 21.8 \\
			CDFI~\cite{ding2021cdfi} & Vimeo90K & \textcolor{blue}{\textbf{35.17}} & 0.964 & \textcolor{blue}{\textbf{35.21}} & 0.950 & \textcolor{blue}{\textbf{37.14}} & 0.966 & \textcolor{red}{\textbf{1.98}} & 0.178 & 5.0 \\
			EDSC~\cite{9501506} & Vimeo90K & 34.84 & \textcolor{blue}{\textbf{0.975}} & 35.13 & \textcolor{blue}{\textbf{0.968}} & 36.76 & 0.966 & 2.02 & 0.043 & 8.9 \\
			EA-Net~\cite{9791082} & Vimeo90K & 34.39 & \textcolor{blue}{\textbf{0.975}} & 34.97 & \textcolor{blue}{\textbf{0.968}} & - & - & - & - & - \\
			PMCRNet (Ours) & Vimeo90K & \textcolor{red}{\textbf{35.76}} & \textcolor{red}{\textbf{0.979}} & \textcolor{red}{\textbf{35.29}} & \textcolor{red}{\textbf{0.969}} & \textcolor{red}{\textbf{37.35}} & \textcolor{red}{\textbf{0.985}} & \textcolor{blue}{\textbf{2.00}} & \textcolor{red}{\textbf{0.016}} & 6.2 \\
			\bottomrule
	\end{tabular}}
\end{table*}

Moreover, instead of predicting target texture directly from deep feature like most other VFI algorithms~\cite{8578281,Lee_2020_CVPR,Gui_2020_CVPR,Niklaus_2020_CVPR,BMBC,reda2022film}, we propose a new context synthesis module (CSM) to simplify the frame synthesis in each level by borrowing existing texture from adjacent input image pyramids, which means that decoder $\mathcal{D}^{l}$ of PMCRNet only needs to predict easier intermediate optical flow $F_{t\rightarrow0}^{l-1}, F_{t\rightarrow1}^{l-1}$, one-channel occlusion merge mask $M^{l-1}$ and three-channel image residual $R^{l-1}$ as shown in Fig.~\ref{fig:1}. Note that height and width of tensors with superscript $l$ are $1/2^{l}$ of original input frames. Fig.~\ref{fig:2} gives structure details of the CSM in decoder $\mathcal{D}^{l}$ ($l \in \{1, 2, 3, 4\}$), which can be formulated as
\begin{gather}
	\hat{I}_{t}^{l-1} = M^{l-1} \otimes \tilde{I}_{0}^{l-1} + (1 - M^{l-1}) \otimes \tilde{I}_{1}^{l-1} + R^{l-1}, \\
	\tilde{I}_{0}^{l-1} = w(I_{0}^{l-1}, F_{t\rightarrow0}^{l-1}), \, \tilde{I}_{1}^{l-1} = w(I_{1}^{l-1}, F_{t\rightarrow1}^{l-1}),
\end{gather}
where $\otimes$ means element-wise multiplication. $I_{0}^{l-1}, I_{1}^{l-1}$ are down-sampled input frames with scale factor $1/2^{l-1}$, and $\hat{I}_{t}^{0}$ is the final predicted target frame. Subsequent experiments will show that proposed CSM can not only generate clear motion texture but also achieve better interpolation accuracy.

\subsection{Loss Function}
Based on above analysis, a motion context joint refinement based PMCRNet has been built for VFI. To optimize this network, we employ the Charbonnier loss~\cite{413553} $\rho$ to replace the $\mathcal{L}_{1}$ loss like many existing methods~\cite{8578281,8954114,Niklaus_2020_CVPR,BMBC,reda2022film}. Besides, inspired by the robustness of the census loss $\mathcal{L}_{cen}$ in unsupervised optical flow estimation~\cite{Meister_2018,9882137}, we add it as a complementary loss term, which calculates the soft Hamming distance between census-transformed image patches of size 7$\times$7. Formally, our image reconstruction loss $\mathcal{L}_{r}$ can be denoted as
\begin{equation}
	\mathcal{L}_{r}(\hat{I}_{t}^{0}, I_{t}^{0}) = \rho(\hat{I}_{t}^{0} - I_{t}^{0}) + \mathcal{L}_{cen}(\hat{I}_{t}^{0}, I_{t}^{0}),
\end{equation}
where $\rho(x) = (x^2 + \epsilon^2)^{\alpha}$ with $\alpha = 0.5, \epsilon = 10^{-3}$ is the Charbonnier loss, $\hat{I}_{t}^{0}, I_{t}^{0}$ are predicted and ground truth frame.

Above objective can already supervise frame interpolation, however, it often drops into local minimum due to misaligned texture in large motion cases. To solve this problem, DAIN~\cite{8954114}, SoftSplat~\cite{Niklaus_2020_CVPR} use additional pre-trained optical flow network, RIFE~\cite{huang2021rife}, IFRNet~\cite{Kong_2022_CVPR} leverage knowledge distillation from the teacher flow network. However, cascaded architecture and knowledge distillation can result in large inference delay and complicate training process. In this letter, we propose a new annealed multi-scale reconstruction loss to better guide the learning procedure. Our motivation is based on the observation that multi-scale reconstruction is helpful in the early stage of training which can help to jump out from local minimum, while it can impede final accuracy in the late stage of optimization. Therefore, our final loss can be formulized as
\begin{equation}
	\mathcal{L} = \mathcal{L}_{r}(\hat{I}_{t}^{0}, I_{t}^{0}) + \tau \sum_{l=1}^{3} \mathcal{L}_{r}(\hat{I}_{t}^{l}, I_{t}^{l}),
	\label{eq:6}
\end{equation}
where $\tau$ is the weighting coefficient. In our experiment, $\tau$ decays from 0.1 to 0.0 linearly during the first $1/4$ training epochs, and maintains 0.0 in the rest of training epochs.

\section{Experiments}
In this section, we first introduce the implementation details of our PMCRNet. Then, quantitative and qualitative comparisons with other SOTA methods are performed. Finally, we carry out ablation study to verify proposed approaches.

\subsection{Implementation Details}
\subsubsection{Datasets}
We use the training set of Vimeo90K triplet dataset~\cite{xue2019video} to train proposed PMCRNet from scratch and use Vimeo90K triplet test dataset~\cite{xue2019video}, UCF101 test dataset~\cite{Soomro_2012} in DVF~\cite{8237740} and Middlebury-Other dataset~\cite{Baker_2011} for evaluation.

\subsubsection{Metrics}
For all test datasets, we use the common Peak Signal-to-Noise Ratio (PSNR) and Structural Similarity (SSIM)~\cite{1284395} metrics. For Middlebury, the official metric of Interpolation Error (IE) is also adopted.

\subsubsection{Details}
We implement proposed PMCRNet in PyTorch, and use AdamW~\cite{Loshchilov_2019} algorithm to optimize it for 300 epochs with total batch size 16 on four NVIDIA RTX 2080 Ti GPUs. The learning rate is initially set to 1e-4, and gradually decays to 2e-5 according to a cosine attenuation schedule. Data augmentation such as random flipping, rotating, reversing sequence order and random cropping patches with size 256$\times$256 are employed during training. For inference time in Table~\ref{tab:1}, we run the method under 640$\times$480 input resolution on one RTX 2080 Ti GPU, and average the time with 100 iterations.

\begin{figure*}[t]
	\centering
	\resizebox{1.00\textwidth}{!}{
		\begin{tabular}{@{}c @{\hskip 0.02in} c @{\hskip 0.02in} c @{\hskip 0.02in} c @{\hskip 0.02in} c @{\hskip 0.02in} c @{\hskip 0.02in} c @{\hskip 0.02in} c@{}}
			\includegraphics[width=0.195\linewidth]{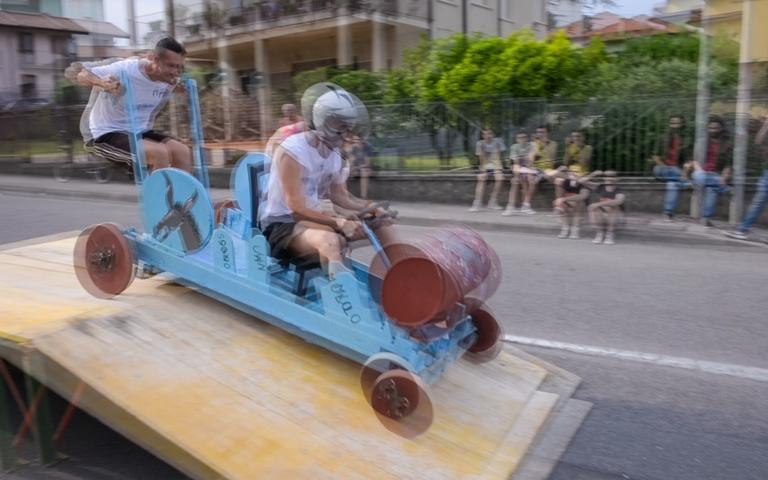}
			&
			\includegraphics[width=0.195\linewidth]{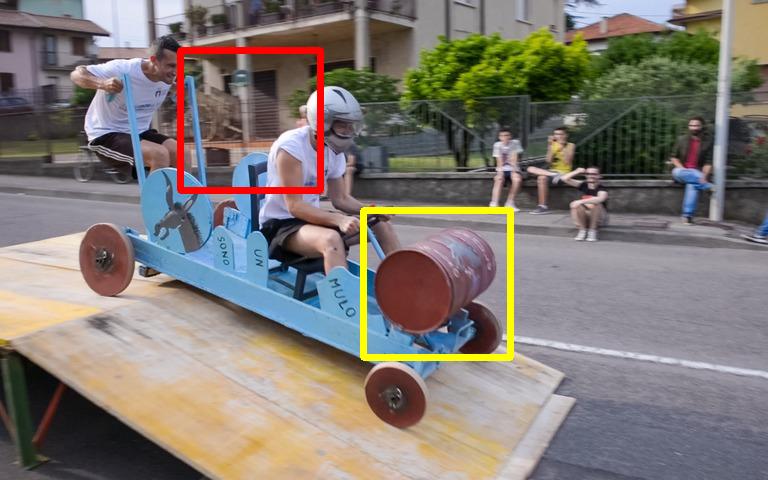}
			&
			\includegraphics[width=0.195\linewidth]{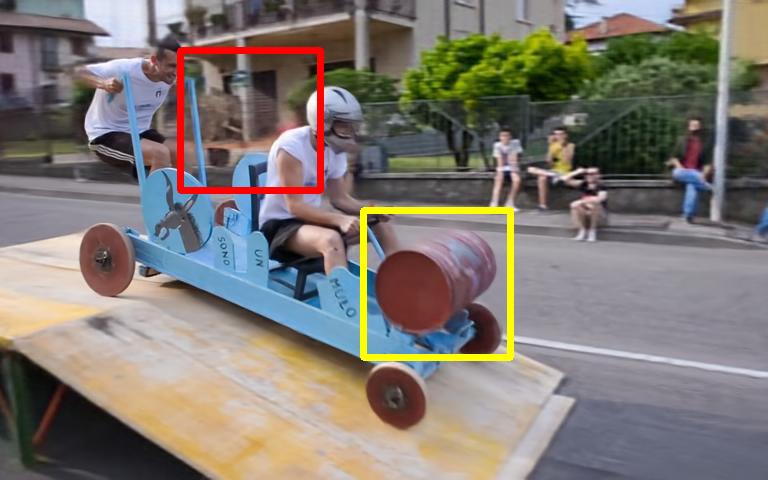}
			&
			\includegraphics[width=0.195\linewidth]{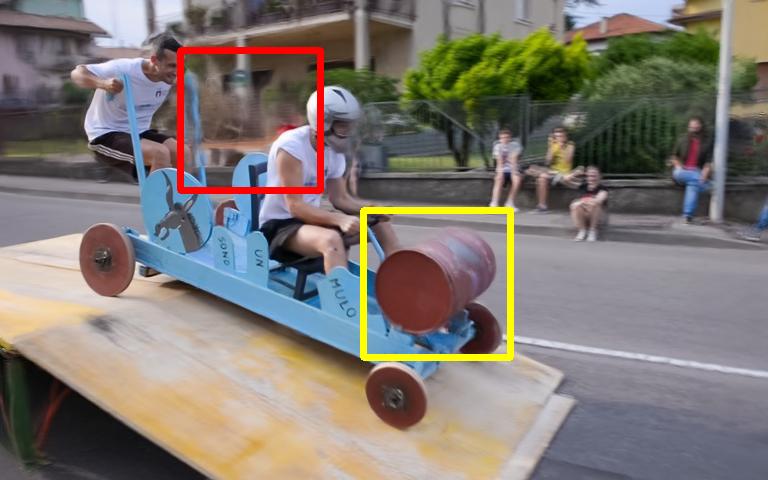}
			&
			\includegraphics[width=0.195\linewidth]{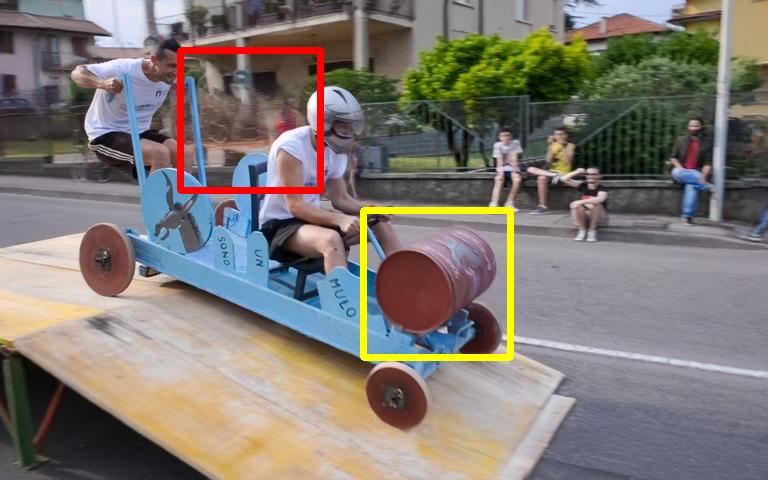}
			&
			\includegraphics[width=0.195\linewidth]{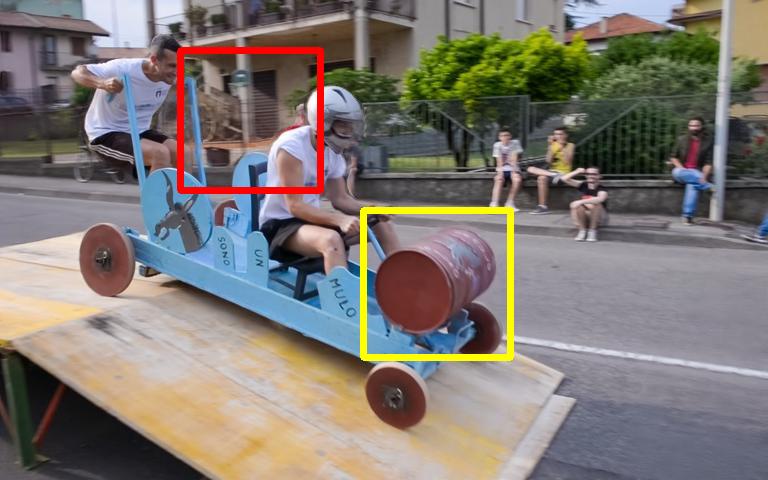}
			\vspace{-0.5mm}
			\\
			\includegraphics[width=0.195\linewidth]{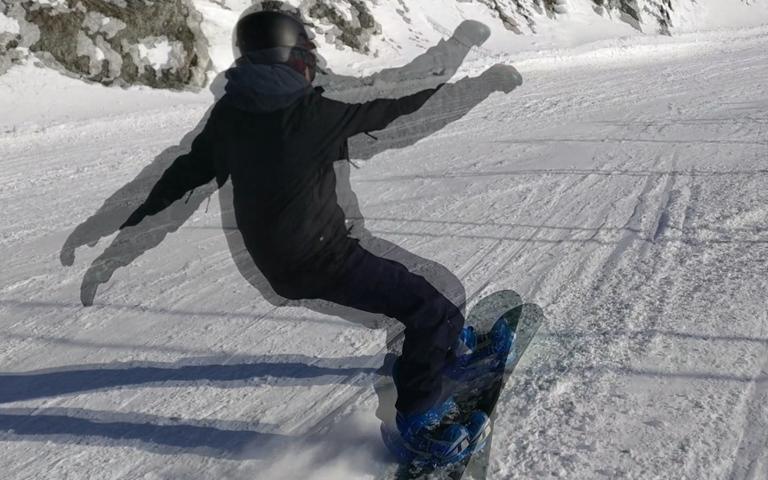}
			&
			\includegraphics[width=0.195\linewidth]{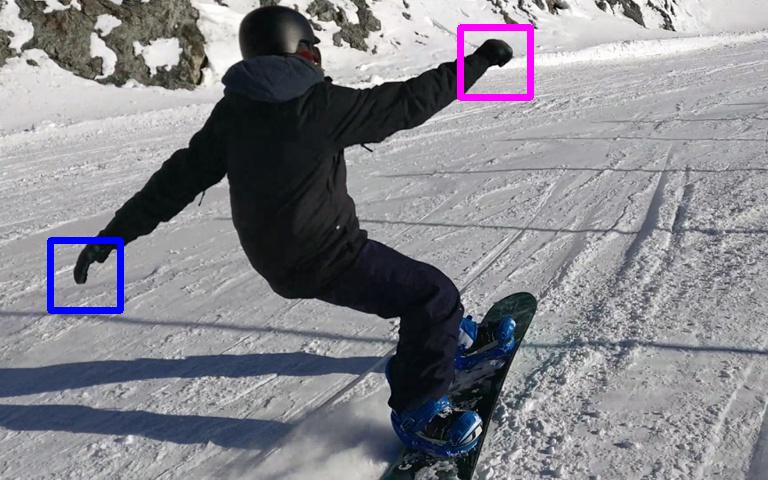}
			&
			\includegraphics[width=0.195\linewidth]{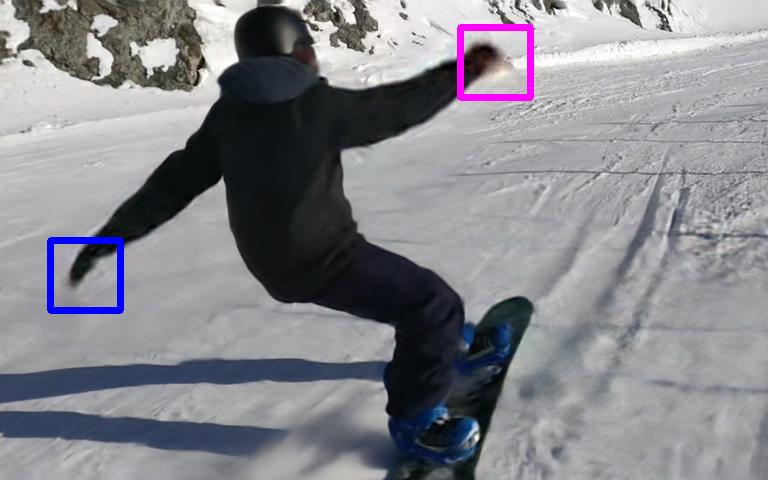}
			&
			\includegraphics[width=0.195\linewidth]{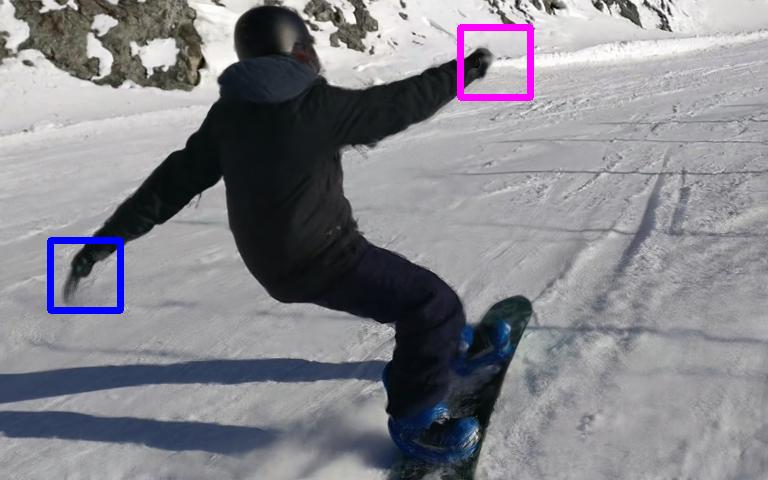}
			&
			\includegraphics[width=0.195\linewidth]{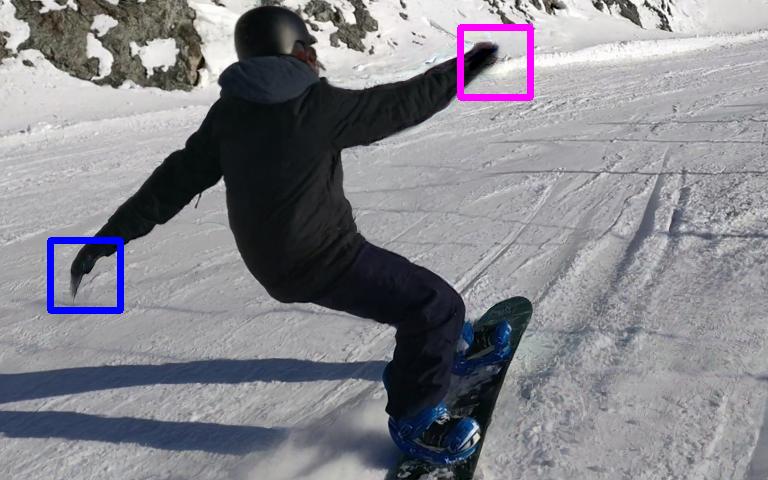}
			&
			\includegraphics[width=0.195\linewidth]{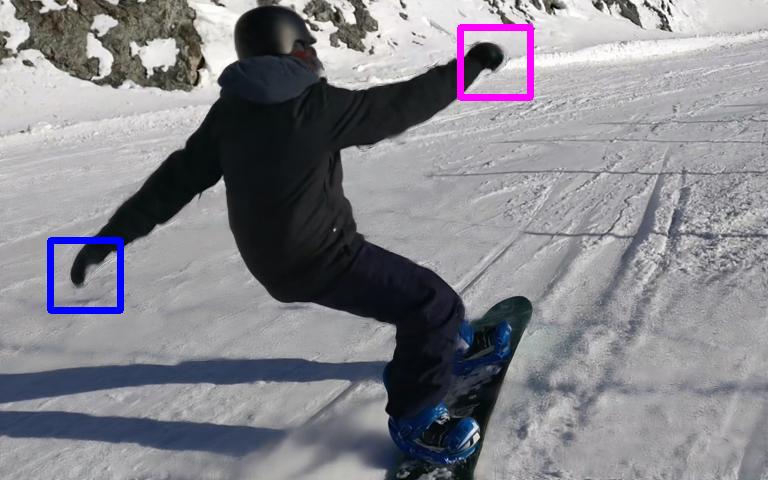}
			\vspace{-0.5mm}
			\\
			Input Overlay & Ground Truth & CAIN~\cite{choi2020cain} & AdaCoF~\cite{Lee_2020_CVPR} & CDFI~\cite{ding2021cdfi} & PMCRNet (Ours)
	\end{tabular}}
	\caption{Qualitative comparison with state-of-the-art VFI methods that do not use additional flow knowledge on DAVIS dataset~\cite{Jordi_arXiv_2017}. Zoom in for best view.}
	\label{fig:3}
\end{figure*}

\subsection{Comparison to the State-of-the-Arts}
In this part, we compare proposed PMCRNet with state-of-the-art VFI methods quantitatively and qualitatively. Note that only methods which do not use additional optical flow information are listed in Table~\ref{tab:1} for fair comparison. For PSNR on Vimeo90K and UCF101 test datasets, proposed PMCRNet outperforms SOTA method CDFI~\cite{ding2021cdfi} by 0.59 dB and 0.08 dB, respectively. Moreover, our motion context joint refinement based network runs 11$\times$ faster than multi-branch based CDFI~\cite{ding2021cdfi}, demonstrating the efficiency of our approach. For SSIM, our PMCRNet behaves better than recent methods EDSC~\cite{9501506} and EA-Net~\cite{9791082} on Vimeo90K, and achieves the same best accuracy as CAIN~\cite{choi2020cain} on UCF101 but containing 6.9$\times$ less parameters. As for Middlebury benchmark, we also rank 1st on both PSNR and SSIM, while only fall behind CDFI~\cite{ding2021cdfi} by 0.02 on the IE metric. Fig.~\ref{fig:3} qualitatively compares our PMCRNet with several advanced methods in two large motion cases on DAVIS dataset~\cite{Jordi_arXiv_2017}. As highlighted by the bounding boxes, our approach can generate more faithful high-frequency texture details in the top example and synthesize realistic motion boundaries in the bottom example.

\begin{table}[t]
	\centering
	\renewcommand{\arraystretch}{0.9}
	\caption{Ablation study on proposed architecture and loss function. `PMR' and `PCR' stand for progressive motion refinement and progressive context refinement in our PMCRNet. $\tau$ is the weighting coefficient in Eq.~\ref{eq:6}. $\rightarrow$ means the annealing process.}
	\vspace{-1.5mm}
	\label{tab:2}
	{\footnotesize
		\setlength\tabcolsep{8.4pt}
		\begin{tabular}{ccccc}
			\toprule
			ID & Structure & Loss & Vimeo90K & Middlebury \\
			\midrule
			E1 & w/o PMR & $\tau: 0.0$ & 34.66 & 35.93 \\
			E2 & w/o PCR & $\tau: 0.0$ & 35.41 & 36.88 \\
			E3 & w/o CSM & $\tau: 0.0$ & 35.19 & 36.25 \\
			\midrule
			E4 & PMCRNet & $\tau: 0.0$ & 35.61 & 36.98 \\
			E5 & PMCRNet & $\tau: 0.1$ & 35.47 & 36.74 \\
			E6 & PMCRNet & $\tau: 0.1\rightarrow0.0$ & \textbf{35.76} & \textbf{37.35} \\
			\bottomrule
	\end{tabular}}
	\vspace{-2.0mm}
\end{table}

\begin{figure}[t]
	\centering
	\scriptsize
	\resizebox{0.495\textwidth}{!}{
		\begin{tabular}{@{}c @{\hskip 0.02in} c @{\hskip 0.02in} c @{\hskip 0.02in} c @{\hskip 0.02in} c @{\hskip 0.02in} c@{}}
			\includegraphics[width=0.196\linewidth]{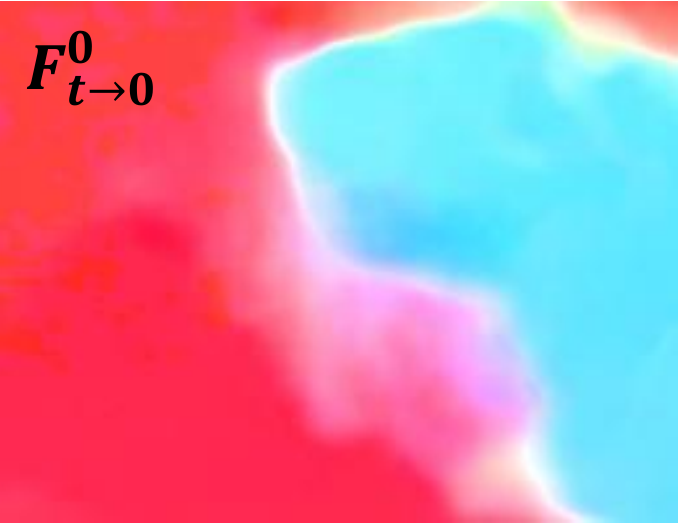}
			&
			\includegraphics[width=0.196\linewidth]{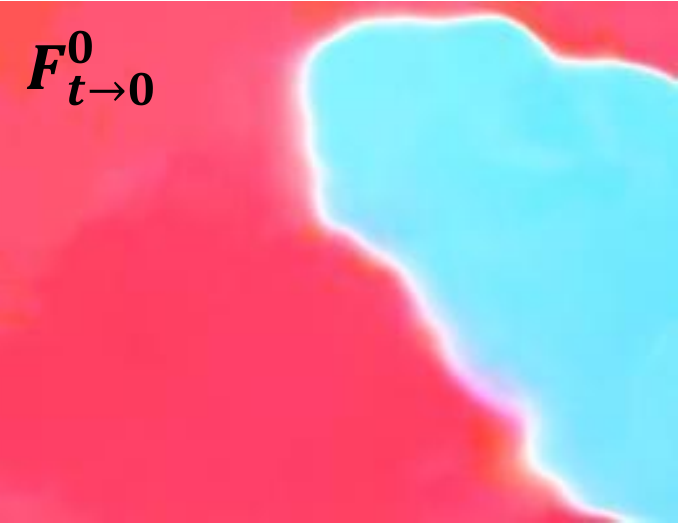}
			&
			\includegraphics[width=0.196\linewidth]{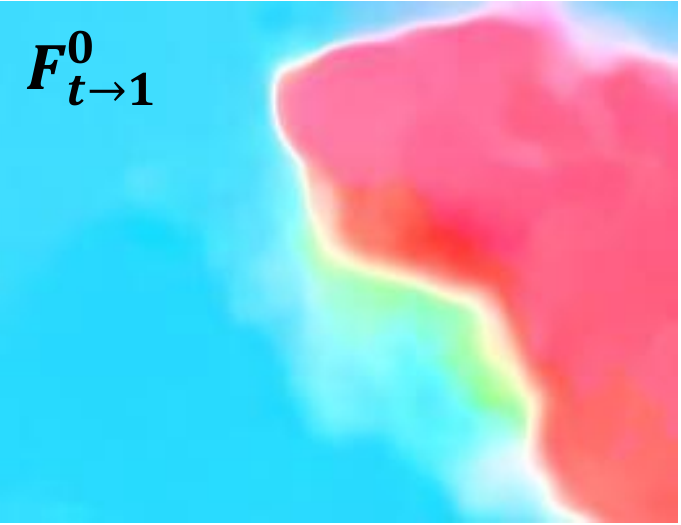}
			&
			\includegraphics[width=0.196\linewidth]{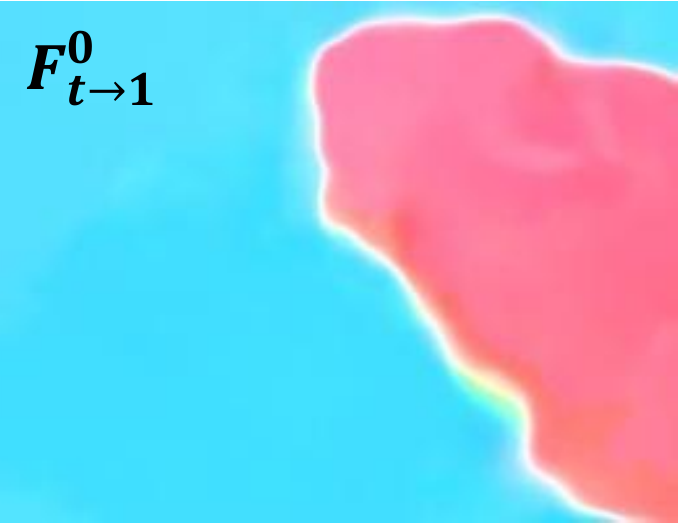}
			\vspace{-0.5mm}
			\\
			w/o AMSR (E4) & Full (E6) & w/o AMSR (E4) & Full (E6)
			\\
			\includegraphics[width=0.196\linewidth]{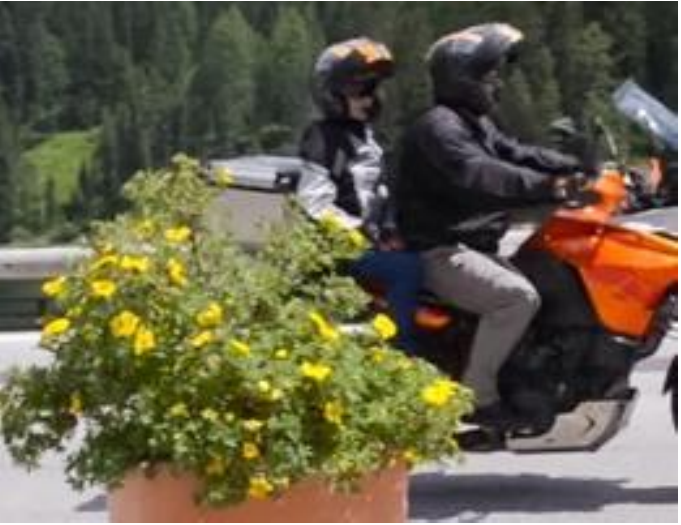}
			&
			\includegraphics[width=0.196\linewidth]{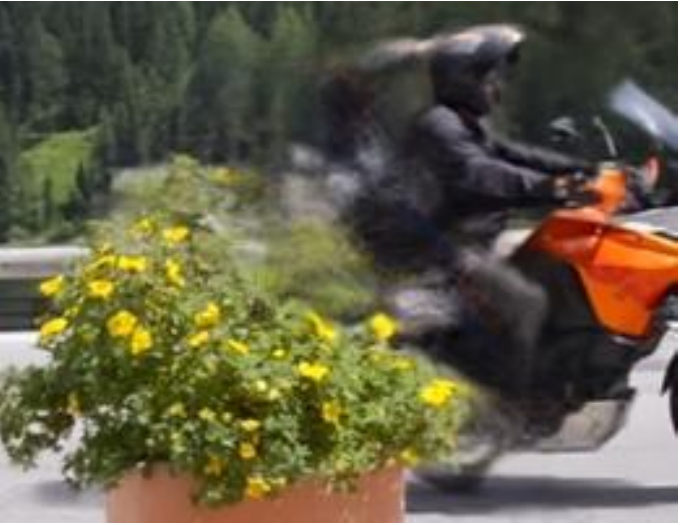}
			&
			\includegraphics[width=0.196\linewidth]{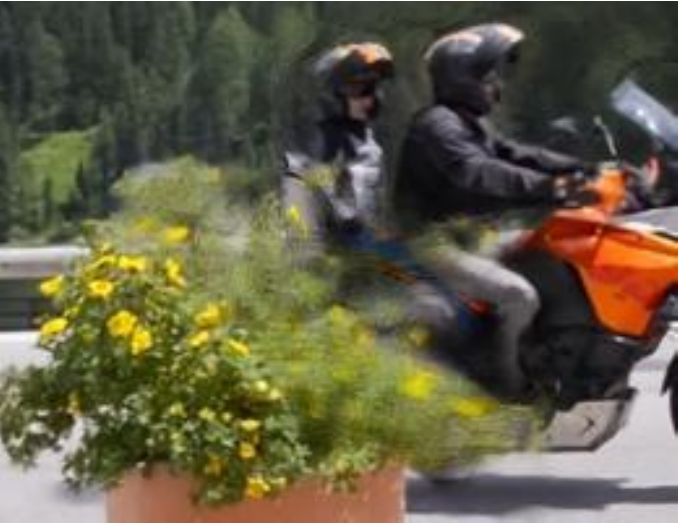}
			&
			\includegraphics[width=0.196\linewidth]{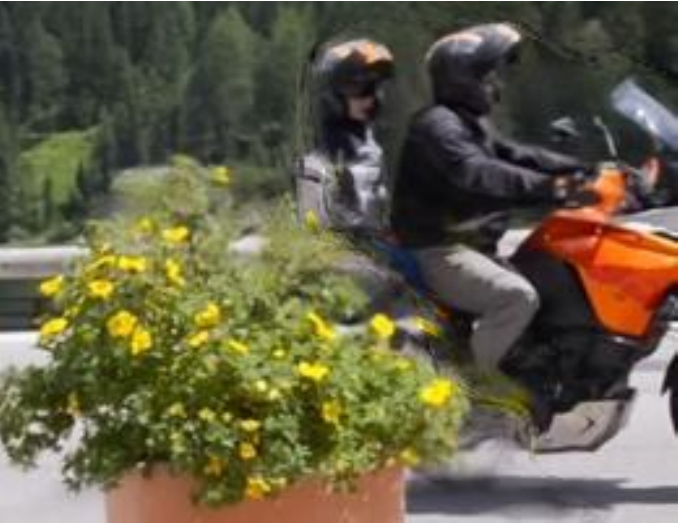}
			\vspace{-0.5mm}
			\\
			Ground Truth & w/o CSM (E3) & w/o AMSR (E4) & Full (E6)
	\end{tabular}}
	\caption{Qualitative comparison among different ablations. `AMSR' means proposed annealed multi-scale reconstruction loss. The first row is visualized intermediate flow of $F_{t\rightarrow0}^{0}$ and $F_{t\rightarrow1}^{0}$. The second row is predicted frame.}
	\label{fig:4}
\end{figure}

\subsection{Ablation Study}
To verify the effectiveness of our contributions, we carry out ablation experiments on proposed VFI architecture and loss function, whose quantitative and qualitative results are shown in Table~\ref{tab:2} and Fig.~\ref{fig:4} separately.

\subsubsection{Network Structure}
E1, E2, E3 in Table~\ref{tab:2} represent three network variants by removing progressive motion refinement (PMR), progressive context refinement (PCR) and context synthesis module (CSM) from proposed PMCRNet respectively. E4 is the whole PMCRNet for reference. E1, E2, E3 and E4 are all optimized by the same image reconstruction loss of Eq.~\ref{eq:6} with $\tau=0.0$ for fair comparison. Specifically, E1 removes $F_{t\rightarrow0}^{l}, F_{t\rightarrow1}^{l}, M^{l}, R^{l}$ and replaces $\tilde{\phi}_{0}^{l}, \tilde{\phi}_{1}^{l}$ by $\phi_{0}^{l}, \phi_{1}^{l}$ in Eq.~\ref{eq:2}, which means to remove previous level motion but keep previous level context information. On the contrary, E2 removes previous level context information $\hat{I}_{t}^{l}$ but keeps previous level motion information. It can be seen that E1 drops by about 1 dB on both benchmarks from E4, while E2 only drops 0.2 dB and 0.1 dB from E4 on Vimeo90K and Middlebury. Therefore, progressive motion refinement and progressive context refinement are both helpful and complementary for proposed PMCRNet to achieve advanced VFI accuracy, while motion refinement is more important than context refinement due to the large displacement challenge in frame interpolation. E3 is the ablation which replaces proposed context synthesis module by directly predicting context information from deep feature. Compare E3 with E4, we can see an obvious reduction on accuracy in Table~\ref{tab:2}, and the predicted frame becomes blurrier in Fig.~\ref{fig:4}, e.g., the man on the back of the motorcycle.

\subsubsection{Loss Function}
E4, E5 and E6 in Table~\ref{tab:2} stand for without multi-scale reconstruction loss, fixed multi-scale reconstruction loss and annealed multi-scale reconstruction loss when optimizing PMCRNet. We can observe that E5 behaves even worse than E4, it is because that fixed multi-scale reconstruction can impede final accuracy in the late stage of training. When we gradually anneal $\tau$ from 0.1 to 0.0, the final accuracy improves. The reason is that multi-scale reconstruction is helpful in the early stage of training which can help to jump out from local minimum. Compare E4 with E6 in Fig.~\ref{fig:4}, intermediate flow fields $F_{t\rightarrow0}^{0}, F_{t\rightarrow1}^{0}$ of E6 keep clearer motion boundary and behave more certain than those of E4. Correspondingly, more faithful texture is generated on the potted plants of E6, verifying the effectiveness of our annealed multi-scale reconstruction loss. A failure case in Fig.~\ref{fig:4} is that the box behind the person is wrongly synthesized to the tree, because of wrongly estimated flow in complex occlusion cases.

\section{Conclusion}
In this letter, we have presented a novel single encoder-decoder based PMCRNet for efficient video frame interpolation. Different from existing cascaded flow-based VFI deep architectures, our network jointly refines bidirectional optical flow and target frame context in a coarse-to-fine manner that can not only deal with the challenge of large displacement but also reduce model size and inference delay. Moreover, a new context synthesis module is embedded into each decoder that can simplify target frame generation by borrowing existing texture from adjacent input frames and synthesize more faithful high frequency details. Finally, a new annealed multi-scale reconstruction loss is introduced to provide better supervision signal by only imposing regularization constraints in the early stage of training. Experiments on multiple benchmarks show that our approaches can achieve state-of-the-art VFI accuracy while maintaining lightweight and fast.

\bibliographystyle{IEEEtran}
\bibliography{reference}

\end{document}